%% file: bmvc_final.tex
\documentclass{bmvc2k}
\usepackage{graphicx}
\usepackage{booktabs}
\input{math_commands.tex}

\usepackage{hyperref}
\usepackage{url}
\usepackage{graphicx} 
\usepackage{algorithm}
\usepackage{colortbl}
\usepackage{xcolor}
\definecolor{myblue}{RGB}{30, 10, 255} 
\definecolor{mygreen}{RGB}{223, 252, 220} 
\usepackage{algorithmic}
\usepackage{booktabs}
\usepackage{multirow}
\usepackage{caption}
\usepackage{hyperref}

\usepackage{orcidlink}

\usepackage[accsupp]{axessibility}  

\title{Beyond Global Editing: Per-Instance Disentangled Subspaces for Training-Free Hallucination Mitigation in LVLMs}

\addauthor{Ali Cheraghian}{ali.cheraghian@mq.edu.au}{1, 7}
\addauthor{Hamidreza Dastmalchi}{hrd@yorku.ca}{2}
\addauthor{Hamed Barzamini}{h.barzamini@niu.edu}{3}
\addauthor{Morteza Saberi}{Morteza.Saberi@uts.edu.au}{4}
\addauthor{Mojtaba Golzan}{Mojtaba.Golzan@uts.edu.au}{4}
\addauthor{Shafin Rahman}{shafin.rahman@northsouth.edu}{5}
\addauthor{Hossein Rahmani}{h.rahmani@lancaster.ac.uk}{6}

\addinstitution{
Macquarie University, Australia
}
\addinstitution{
 York University, Canada
}

\addinstitution{
Northern Illinois University, USA
}

\addinstitution{
University of Technology Sydney, \\ Australia
}

\addinstitution{
North South University, Bangladesh
}

\addinstitution{
Lancaster University, UK
}

\addinstitution{
Australian National University, Australia
}

\runninghead{Cheraghian et al}{Disentangled Subspaces for LVLMs}

\begin{document}

\maketitle

\begin{abstract}
 Recent advances in large vision-language models (LVLMs) have enabled powerful multimodal reasoning by integrating visual encoders with large language models (LLMs). However, their reliability is frequently undermined by hallucinations, where generated text inaccurately describes the visual input. Although fine-tuning can mitigate this problem, it is computationally expensive and requires large, curated datasets, making training-free alternatives attractive. Among these, model editing is more promising than decoding-based approaches: decoding methods adapt outputs per input but introduce computational overhead and instability, whereas model editing modifies internal representations offline, providing a more efficient and stable solution. However, existing model-editing techniques typically rely on a single global subspace to correct hallucinations, treating all test samples identically and failing to capture diverse hallucination modes across inputs. To address this limitation, we propose a training-free hallucination mitigation framework for dynamic, per-instance suppression at test time. Our method first constructs a set of Disentangled Hallucination Subspaces, each isolating a distinct hallucination mode. During inference, the model adaptively calculates weights reflecting each input's relationship to these subspaces, guiding a dynamically combined projection that selectively suppresses the most probable hallucination directions while preserving image-grounded semantics. Extensive experiments across multiple vision-language benchmarks and LVLM families demonstrate consistent improvements, highlighting the robustness, generalizability, and efficiency of our approach.

\end{abstract}

\section{Introduction}
\label{sec:intro}

In recent years, integrating vision models with large language models (LLMs) has become a standard approach to leverage the advanced reasoning capabilities of LLMs. This integration has led to the emergence of large vision-language models (LVLMs)~\cite{liu2023llava, zhu2023minigpt}. Despite rapid progress in this field, LVLMs remain unreliable in certain scenarios due to the persistent issue of hallucination, where the model generate irrelevant or non-factual content that is inconsistent with the input image.

\begin{figure}[t]
    \centering
    \includegraphics[width=\linewidth]{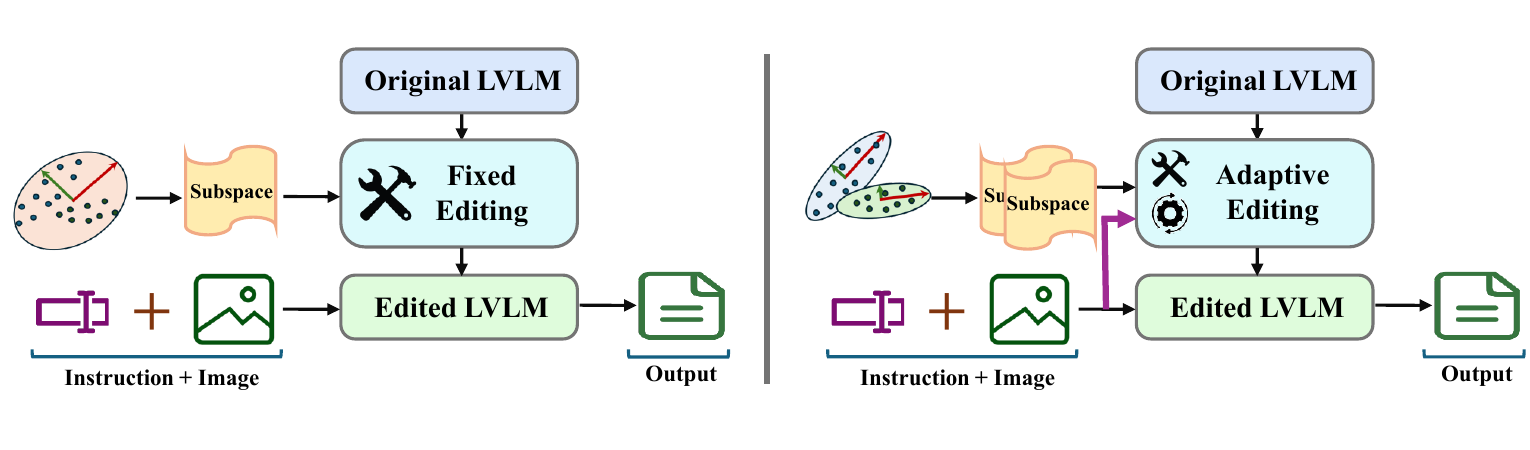}
    \vspace{0mm}
    \caption{
    \textbf{(Left)}: Existing model editing methods ~\cite{yang2025nullu} derive a single subspace of hallucination directions (via SVD) and apply fixed editing to the LVLM, using the same edited model for all inputs. \textbf{(Right)}: Our adaptive method identifies multiple subspaces from different hallucination modes and adaptively adjusts their contributions to edit the model based on the input image, enabling more flexible and context-aware hallucination suppression.
    }
    \label{fig:motivation}

\end{figure}

Existing strategies for reducing hallucinations in LVLMs generally fall into two main categories: (i) fine-tuning approaches~\cite{xiao2025detecting, yu2024rlhf} and (ii) training-free methods~\cite{leng2024mitigating, wang-etal-2024-mitigating, yang2025nullu}. Although the fine-tuning approach often achieves superior performance, it requires curated datasets and significant computational resources, limiting its practicality in real-world deployments. Thus, training-free methods have become increasingly popular due to their flexibility and efficiency.

Existing training-free techniques can be broadly categorized into decoding-based~\cite{leng2024mitigating, wang-etal-2024-mitigating} and model-editing~\cite{yang2025nullu} approaches. Decoding-based methods, such as contrastive decoding~\cite{leng2024mitigatingvcd}, adapt model outputs per input during or after generation. Although effective, they incur substantial computational overhead due to additional forward passes and often suffer from instability. In contrast, model-editing approaches modify the internal layers of LVLMs offline, typically by projecting hidden representations into a low-dimensional subspace that captures hallucination-sensitive directions. 
However, prior methods rely on a single global subspace that is applied uniformly across all test samples, which limits their generalization capacity: the hallucination direction captured by this subspace may not align with the diverse hallucination patterns observed during inference. Specifically, existing editing approaches~\cite{yang2025nullu} (see Figure~\ref{fig:motivation} \textbf{(left)}) perform fixed model editing, intervening in the internal layers solely according to this global subspace, regardless of the input. This motivates the need for an adaptive strategy that can dynamically adjust at test time. To this end, we propose an adaptive model-editing framework (see Figure~\ref{fig:motivation} \textbf{(right)}) that leverages multiple disentangled subspaces. Each subspace is better aligned with a particular hallucination mode, and their contributions to model editing are adaptively weighted based on the input sample itself. Overall, our framework operates in two steps:

\textbf{Step 1:} In a preprocessing stage, we construct multiple low-dimensional subspaces, each representing a distinct hallucination direction. Specifically, we leverage a paired dataset where each image is associated with both a hallucinated and a truthful description. For every image, we compute the state differences between hallucinated and truthful captions. These per-layer difference vectors are then clustered using K-means. Within each cluster, we apply Singular Value Decomposition (SVD) to extract an orthonormal basis that captures the dominant hallucination direction. This process yields a collection of subspaces, each characterizing a different hallucination-sensitive direction.

\textbf{Step 2:} At inference time, the precomputed subspaces from the offline stage are adaptively combined based on the given test sample. Specifically, we probe the LVLM with both the original image and a masked variant designed to induce hallucination. The difference between their hidden states provides an input-specific hallucination signal, capturing the model’s susceptibility to hallucinated content. This signal is then projected onto the previously constructed subspaces, where the projection magnitudes act as relevance scores. These scores are used to compute adaptive weights for combining the corresponding subspace bases. The resulting weighted projection matrix is finally applied to the model’s internal representations to suppress hallucinations during inference.

Unlike fixed editing methods~\cite{yang2025nullu}, our framework enables adaptive, fine-grained, and context-aware model editing, effectively filtering hallucinated content while preserving image-grounded semantics.  In summary, our contributions are as follows:

\begin{itemize}

 \item \textbf{Disentangled Hallucination Subspaces:} an offline construction of multiple low-rank subspaces, each capturing a distinct hallucination mode.  
    
\item \textbf{Adaptive Test-Time Mitigation:} a training-free framework that extracts an input-specific hallucination signal and projects it onto the precomputed subspaces. The resulting adaptive weighting dynamically suppresses hallucination directions during inference.

\item \textbf{Extensive Evaluation:} We conduct experiments across multiple vision-language benchmarks and VLM families, demonstrating consistent and generalizable improvements across evaluation metrics.  
\end{itemize}

\section{Related work}

\noindent\textbf{Large Vision-Language Models:} 
Large Vision-Language Models (LVLMs) have advanced rapidly, with designs such as BLIP-2~\cite{li2023blip}, InstructBLIP~\cite{instructblip}, MiniGPT-4~\cite{zhu2023minigpt}, LLaVA~\cite{liu2023visual}, mPLUG-Owl2~\cite{ye2024mplug}, and Qwen-VL~\cite{Qwen-VL} enabling strong multimodal capabilities. BLIP-2~\cite{li2023blip} employs a Query Transformer (Q-Former) to extract a fixed set of informative visual tokens from a frozen vision encoder, which are then passed to a frozen LLM—enabling modular alignment with minimal adaptation. InstructBLIP~\cite{instructblip} builds on BLIP-2 by applying instruction tuning across diverse vision-language tasks to improve generalization and alignment. However, bottlenecks such as the limited number of visual tokens and reliance on frozen backbones can lead to vision-to-language misalignment, contributing to hallucination risks. Linear projection approaches, such as MiniGPT-4~\cite{zhu2023minigpt} and early LLaVA~\cite{liu2023visual}, preserve CLIP visual features but rely on weakly supervised tuning, which risks semantic misalignment during generation. Subsequent LLaVA variants~\cite{akl2026himemitigatingobjecthallucinations} address this by replacing linear projections with MLP connectors and scaling visual instruction data, yielding improved visual grounding and more faithful caption generation. mPLUG-Owl2~\cite{ye2024mplug} improves grounding through adaptive modality collaboration modules and broad instruction tuning across diverse tasks. Hallucinations—where models generate content ungrounded in the image—are exacerbated by incomplete grounding and the use of next-token training objectives that fail to penalize unfaithful outputs~\cite{li2023evaluating, Dastmalchi_2025_BMVC, 10943719, 10.1007/978-3-031-72940-9_3, 10.1007/978-981-96-0972-7_11, 9711372}. This issue is further compounded by the tendency of LLMs to exploit language priors, generating statistically plausible but visually unsupported content when visual signals are ambiguous or under-represented in training distributions. Early work in image captioning attributes hallucination to biased decoders and limited visual grounding, foreshadowing similar challenges in modern LVLMs. Subsequent evaluations~\cite{li2023evaluating} emphasize that next-token training objectives further promote unfaithful outputs due to weak alignment constraints. Benchmark datasets such as CHAIR~\cite{rohrbach2018object} and POPE~\cite{li2023evaluating} have exposed persistent grounding failures, revealing that hallucination is not an isolated artifact but a systemic challenge tied to how visual evidence is encoded, propagated, and attended to throughout the decoding process. To address these issues, LLaVA-RLHF~\cite{2023llavarlhf} introduces factually-augmented reinforcement learning with human feedback (RLHF) and proposes MMHalBench, a benchmark specifically designed to assess hallucination in LVLMs, showing that targeted supervision can significantly improve factual alignment.

\noindent\textbf{Hallucination Mitigation Strategies:} Mitigation strategies for hallucination in vision-language models generally fall into four categories. {(1) Null-space projection:} Nullu~\cite{yang2025nullu} suppresses hallucinations by projecting input features into the null space of a learned hallucination subspace (HalluSpace). While effective, it applies a single global HalluSpace and cannot adapt to sample-specific hallucination patterns. {(2) Latent space steering:} VTI~\cite{liu2024reducing} applies fixed latent offsets to stabilize vision-language features during decoding. Although training-free, it lacks input adaptivity, as fixed shifts may be suboptimal across different scenes or prompts. {(3) Token sparsification and contrastive decoding:} VASparse~\cite{zhuang2025vasparse} filters low-importance visual tokens using attention sparsity and performs contrastive decoding between full and pruned tokens. While efficient, it does not semantically model hallucination features. {(4) Constrained and contrastive decoding:} Several methods fall into this category. HALC~\cite{chen2024halc} reweights decoding using adaptive visual context and contrast signals. DoLa~\cite{chuang2023dola} compares internal representations to detect and suppress hallucinated content. OPERA~\cite{huang2024opera} applies over-trust penalties and retrospection to discourage unsupported generations. VCD~\cite{leng2024mitigating} performs contrastive decoding using perturbed visual inputs, and Woodpecker~\cite{yin2024woodpecker, Dastmalchi_2026_CVPR} verifies and replaces hallucinated entities by cross-checking them with image-grounded evidence. Despite these advances, most existing methods are either global, heuristic-driven, or computationally intensive. In contrast, our method constructs multiple low-rank Disentangled Hallucination Spaces and performs sample-specific null-space projection at test time. It is entirely training-free, adapts to individual inputs, and avoids reranking or model retraining.

\section{Preliminary}

\noindent\textbf{Vision-Language Alignment.} The input to a vision-language foundation model (LVLM) consists of an image \( \mathbf{I}^{(i)} \in \mathbb{R}^{H \times W \times C} \) and a textual query \( \mathbf{q}^{(i)} \). A vision encoder (e.g., ViT~\cite{dosovitskiy2021an}, CLIP~\cite{pmlr-v139-radford21a}) first extracts image features from \( \mathbf{I}^{(i)} \). These features are then mapped into the language model's input space by a vision-language alignment module (e.g., Q-Former~\cite{10.5555/3618408.3619222} or a linear projection), producing a sequence of \( N \) visual tokens:
\(
\mathbf{X}^{(i)} = [\mathbf{x}^{(i)}_0, \mathbf{x}^{(i)}_1, \dots, \mathbf{x}^{(i)}_{N-1}], \quad \mathbf{x}^{(i)}_n \in \mathbb{R}^d.
\)
Simultaneously, the textual query \( \mathbf{q}^{(i)} \) is tokenized into a sequence of \( M \) tokens:
\(
\mathbf{T}^{(i)} = [\mathbf{t}^{(i)}_N, \mathbf{t}^{(i)}_{N+1}, \dots, \mathbf{t}^{(i)}_{N+M-1}], \quad \mathbf{t}^{(i)}_m \in \mathbb{R}^d.
\)
The combined input to the LVLM is the concatenated sequence \( [\mathbf{X}^{(i)}, \mathbf{T}^{(i)}] \) of total length \( J = N + M \).

\vspace{0.2cm}
\noindent\textbf{Model Forwarding.} The combined input sequence \( [\mathbf{X}^{(i)}, \mathbf{T}^{(i)}] \in \mathbb{R}^{J \times d} \), where \( J = N + M \), is then passed through the language model component of the LVLM. Let \( L \) denote the total number of transformer layers, and \( \mathbf{z}_{\ell,j}^{(i)} \in \mathbb{R}^d \) represent the hidden state corresponding to token index \( j \) at layer \( \ell \) for sample \( i \). The model produces a sequence of contextualized embeddings:
\begin{equation}
\left\{ \mathbf{z}_{\ell,j}^{(i)} \right\}_{\ell=1, j=1}^{L,\, J} = f_\theta^{\text{LVLM}}\left( \mathbf{I}^{(i)}, \mathbf{q}^{(i)} \right).    
\end{equation}

These hidden states serve as the basis for downstream reasoning and generation tasks and are used in subsequent modules for hallucination suppression.

\vspace{0.2cm}
\noindent\textbf{Response Generation.} 
Following the forward pass through the LVLM, the final-layer hidden states \( \{\mathbf{z}_{L,j}^{(i)}\}_{j=1}^{J} \) are used to generate the output response. Specifically, the model performs autoregressive decoding based on the attended multimodal context, producing the textual response token by token. The probability of the next token \( y_{t+1}^{(i)} \) is modeled as:
\begin{equation}
    P\left(y_{t+1}^{(i)} \mid y_{1:t}^{(i)}, \mathbf{z}_{L,1:J}^{(i)}\right) = \text{softmax}\left( \mathbf{W}_o \, \mathbf{h}_t^{(i)} \right),
\end{equation}
where \( \mathbf{h}_t^{(i)} \) is the decoder's hidden state at time \( t \), and \( \mathbf{W}_o \in \mathbb{R}^{V \times d} \) is the output projection matrix for a vocabulary of size \( V \). Decoding continues until an end-of-sequence token is generated or a predefined maximum length is reached.

\section{Method}

Large vision-language models (LVLMs) are highly prone to hallucination, producing textual outputs that are inconsistent with the input image. Existing training-free mitigation methods rely on either sample-agnostic preprocessing or unstable post-hoc heuristics, both of which ignore a crucial property of the phenomenon: hallucination patterns exhibit substantial sample-level heterogeneity, and therefore demand adaptive test-time correction. Two observations sharpen this requirement. First, hallucinations are not monolithic; object-, attribute-, and relation-level errors occupy distinct, only partially overlapping directions in the model's activation space, so collapsing them into a single global correction vector inevitably under-corrects on some samples and over-corrects on others. Second, the appropriate mixture of these directions is itself input-dependent, implying that any correction operator should be \emph{conditioned} on the test instance rather than fixed offline. Motivated by these observations, we introduce a test-time hallucination mitigation framework that (i) constructs multiple low-rank disentangled hallucination subspaces and (ii) performs input-adaptive projection into these subspaces, without fine-tuning and without compromising generation stability. The framework is realised as a two-stage pipeline: a small library of disentangled hallucination subspaces is first discovered from a contrastive corpus, and at inference time these subspaces are composed into a sample-specific projection applied directly to the LVLM's internal hidden states.


\subsection{Contrastive Dataset Construction}

To extract contrastive signals indicative of hallucinations, we construct an offline dataset of paired vision-language inputs in which each triplet consists of an image $\mathbf{I}^{(i)}$, a faithful caption $\mathbf{q}^{(i)}$, and a semantically inconsistent (hallucinated) caption $\tilde{\mathbf{q}}^{(i)}$ describing the same image:
\begin{equation}
\mathcal{D} = \left\{ \left( \mathbf{I}^{(i)}, \mathbf{q}^{(i)}, \tilde{\mathbf{q}}^{(i)} \right) \right\}_{i=1}^{B}.
\end{equation}
The hallucinated captions are obtained by prompting a strong language model (e.g., GPT-4) to introduce plausible but absent objects or distort spatial relationships in the faithful description. Using a strong language model as a perturbation oracle is deliberate: the resulting captions remain fluent, contextually plausible, and lexically close to their faithful counterparts, which isolates the semantic mismatch as the dominant source of representational shift between $\mathbf{q}^{(i)}$ and $\tilde{\mathbf{q}}^{(i)}$. To prevent the recovered subspaces from collapsing onto a single failure mode, the generation prompt is designed to span the principal categories of multimodal hallucination reported in the literature---non-existent objects, incorrect attributes (e.g., colour, shape, count), and erroneous spatial or relational descriptions. This diversity is essential for the subsequent clustering step, which relies on it to recover semantically distinct hallucination directions.

To analyze the latent representations induced by truthful versus hallucinated captions, we pass each pair through the LVLM and collect hidden states across all layers and positions:
\begin{align}
\left\{ \mathbf{z}_{\ell,j}^{(i)} \right\}_{\ell,j=1}^{L,\, J} 
&= f_{\boldsymbol{\theta}}^{\text{LVLM}}\left( \mathbf{I}^{(i)}, \mathbf{q}^{(i)} \right) \\
\left\{ \tilde{\mathbf{z}}_{\ell,j}^{(i)} \right\}_{\ell,j=1}^{L,\, J} 
&= f_{\boldsymbol{\theta}}^{\text{LVLM}}\left( \mathbf{I}^{(i)}, \tilde{\mathbf{q}}^{(i)} \right)
\end{align}
Here, \( \mathbf{z}_{\ell,j}^{(i)} \) and \( \tilde{\mathbf{z}}_{\ell,j}^{(i)} \) represent the hidden state at layer \( \ell \) and position \( j \), corresponding to the faithful and hallucinated captions, respectively. Critically, the image input is held fixed across the two forward passes, so any systematic difference between \( \mathbf{z}_{\ell,j}^{(i)} \) and \( \tilde{\mathbf{z}}_{\ell,j}^{(i)} \) can be attributed to the semantic perturbation in the textual stream rather than to visual variability. This controlled construction is what allows the resulting difference vectors to be interpreted as approximately pure ``hallucination directions'' in the representation space.

\begin{figure*}[t]
    \centering
    \includegraphics[width=\linewidth]{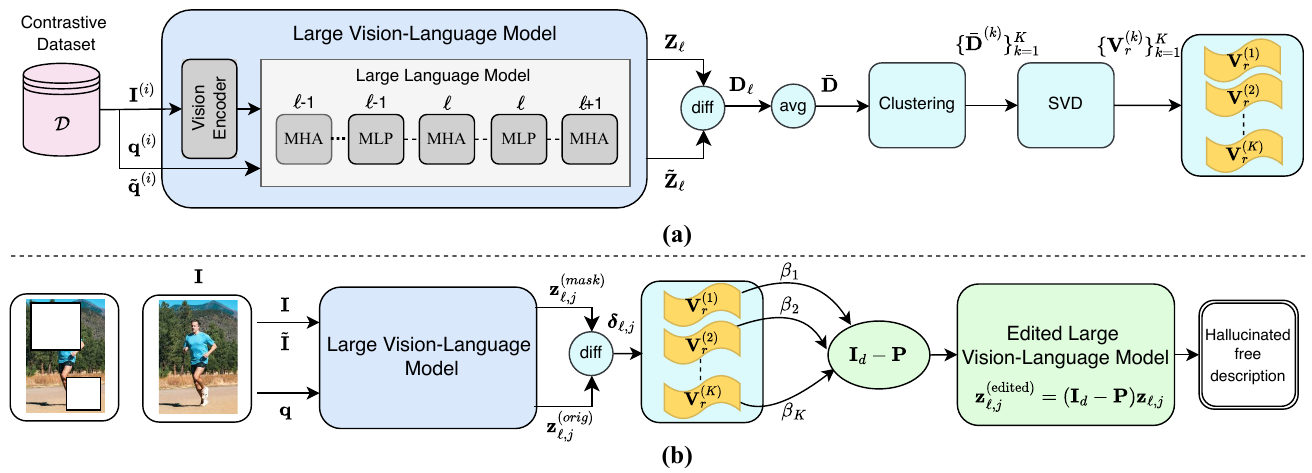}
    \vspace{3mm}
    \caption{ Illustration of our test-time hallucination mitigation framework. \textbf{(a)} We first construct a contrastive dataset~$\mathcal{D}$ to derive a set of low-rank subspaces~$\mathbf{V}^{(i)}_{r}$, $i = 1, \dots, K$, each capturing a distinct type of hallucination. \textbf{(b)} At test time, given an input~$\mathbf{I}$, the model adaptively combines the learned subspaces obtained from clustered hallucination-truthful feature differences—into a sample-specific composite subspace~${\mathbf{P}}$. This subspace is then used to project and modulate internal model activations, effectively suppressing hallucinations without altering the model's parameters.
    }
    \label{fig:test_time_adaptation}
\end{figure*}

\subsection{Disentangled Hallucination Subspaces}

To uncover semantically distinct directions in the LVLM's activation space that correspond to hallucinations, we perform clustering over feature differences derived from the contrastive dataset \( \mathcal{D} \). For each sample \( (\mathbf{I}^{(i)}, \mathbf{q}^{(i)}, \tilde{\mathbf{q}}^{(i)}) \in \mathcal{D} \), we compare the model's internal activations under hallucinated and faithful captions.

Given a set of transformer layers \( \mathcal{L} \), we first compute token-averaged hidden states from each caption. For every layer \( \ell \in \mathcal{L} \), we define:
\begin{equation}
\mathbf{z}_{\ell}^{(i)} = \frac{1}{J} \sum_{j=1}^J \mathbf{z}_{\ell,j}^{(i)}, \quad
\tilde{\mathbf{z}}_{\ell}^{(i)} = \frac{1}{J} \sum_{j=1}^J \tilde{\mathbf{z}}_{\ell,j}^{(i)}
\end{equation}

\noindent where \( \mathbf{z}_{\ell,j}^{(i)} \) and \( \tilde{\mathbf{z}}_{\ell,j}^{(i)} \) are hidden states that correspond to truthful and hallucinated inputs, respectively. It is important to note that token-wise averaging yields a single sequence-level descriptor per layer, which marginalizes positional artifacts arising from differing token lengths between faithful and hallucinated captions, while preserving the layer-specific semantic geometry we wish to characterize.

Inspired by~\cite{yang2025nullu}, we interpret the deviation between these mean activations as a proxy for semantic drift—a key signal of hallucination. For each layer \( \ell \), we collect the representations into matrices:
\begin{align}
\mathbf{Z}_{\ell} &= [\mathbf{z}_{\ell}^{(1)};\, \dots;\, \mathbf{z}_{\ell}^{(B)}] \in \mathbb{R}^{B \times d} \\
\tilde{\mathbf{Z}}_{\ell} &= [\tilde{\mathbf{z}}_{\ell}^{(1)};\, \dots;\, \tilde{\mathbf{z}}_{\ell}^{(B)}] \in \mathbb{R}^{B \times d}
\end{align}

\noindent We compute layer-wise difference matrices:
\begin{equation}
\mathbf{D}_{\ell} = \tilde{\mathbf{Z}}_{\ell} - \mathbf{Z}_{\ell}
\end{equation}

To obtain a consolidated representation of hallucination-induced shifts, we average across all considered layers, $\bar{\mathbf{D}} = \frac{1}{|\mathcal{L}|} \sum_{\ell \in \mathcal{L}} \mathbf{D}_{\ell}$, 

where each row \( \bar{\mathbf{D}}^{(i)} \in \mathbb{R}^d \) now encodes a semantic shift vector for sample \( i \). Averaging across \( \mathcal{L} \) acts as a form of depth-wise denoising: it suppresses layer-specific fluctuations. It retains only the components of the semantic shift that are stable across the depths at which hallucination behaviour is most consistently expressed. We then apply \( K \)-means clustering to the rows of \( \bar{\mathbf{D}} \), identifying groups of hallucination modes:
\begin{equation}
\bar{\mathbf{D}} = [\bar{\mathbf{D}}^{(1)};\, \bar{\mathbf{D}}^{(2)};\, \dots;\, \bar{\mathbf{D}}^{(K)}]
\end{equation}

\noindent where each cluster \( \bar{\mathbf{D}}^{(k)} \in \mathbb{R}^{B_k \times d} \) aggregates samples belonging to hallucination mode \( k \). Clustering before subspace extraction is a key departure from prior single-subspace formulations: it explicitly models hallucinations as a mixture of qualitatively distinct semantic perturbations and prevents the dominant failure mode from absorbing the directions associated with rarer, but equally harmful, error types.

To extract a compact basis for each hallucination mode, we perform Singular Value Decomposition (SVD) on each cluster:
\begin{equation}
\bar{\mathbf{D}}^{(k)} = \mathbf{U}^{(k)} \boldsymbol{\Sigma}^{(k)} {\mathbf{V}^{(k)}}^\top
\end{equation}

\noindent The top-\( r \) right singular vectors from \( \mathbf{V}^{(k)} \), corresponding to the largest singular values, define the basis of the low-rank subspace:
\begin{equation}
\mathbf{V}_r^{(k)} = [\mathbf{v}_{k,1}, \dots, \mathbf{v}_{k,r}] \in \mathbb{R}^{d \times r}
\end{equation}

\noindent Here, \( r \) is a hyperparameter controlling the subspace rank. {By construction, the columns of \( \mathbf{V}_r^{(k)} \) are orthonormal,
\begin{equation}
{\mathbf{V}_r^{(k)}}^\top \mathbf{V}_r^{(k)} = \mathbf{I}_r, \qquad k = 1, \dots, K,
\end{equation}
so that \( \mathbf{V}_r^{(k)} {\mathbf{V}_r^{(k)}}^\top \) is the orthogonal projector onto the \( r \)-dimensional subspace spanned by \( \mathbf{V}_r^{(k)} \). This property is what guarantees that the per-mode projection is idempotent, and that the residual map \( \mathbf{I}_d - \mathbf{V}_r^{(k)} {\mathbf{V}_r^{(k)}}^\top \) acts as the identity on the orthogonal complement of mode~\( k \).} {Truncating to the top-\( r \) singular vectors yields a low-rank summary that captures the principal directions of variation within each hallucination mode while discarding higher-order components that are more likely to reflect sample-specific noise. This low-rank structure is also what makes the subsequent projection-based intervention numerically stable: applying \( \mathbf{I}_d - \mathbf{P} \) removes only a thin, well-conditioned slice of the representation space.}

Finally, we collect all subspaces into the set $\mathcal{V} = \left\{ \mathbf{V}_r^{(k)} \right\}_{k=1}^{K}$.

\noindent Each subspace \( \mathbf{V}_r^{(k)} \) captures a distinct semantic mode of hallucination, enabling fine-grained, projection-based suppression. These subspaces are later used to dynamically remove hallucination directions during inference, enhancing the LVLM's robustness under distribution shift.

\subsection{Test-Time Hallucination Mitigation}

\begin{algorithm}[b!]
\caption{Adaptive Hallucination Suppression via Subspace Projection}
\label{alg:adaptive_projection}
\begin{algorithmic}[1]
\small
\REQUIRE Test image $\mathbf{I}$, query $\mathbf{q}$, vision-language model $f_\theta^{\text{LVLM}}$, hallucination subspaces $\{\mathbf{V}_r^{(k)}\}_{k=1}^K$
\ENSURE Hallucination-suppressed output $y$

\vspace{2pt}
\STATE \textcolor{myblue}{\textbf{\textit{Step 1: Masked Input Generation}}}
\STATE $\mathbf{M} \gets \texttt{Mask}(\mathbf{I}, 0.7)$ \COMMENT{Mask 70\% of image}
\STATE $\tilde{\mathbf{I}} \gets \mathbf{I} \odot \mathbf{M}$

\vspace{2pt}
\STATE \textcolor{myblue}{\textbf{\textit{Step 2: Activation Difference}}}
\STATE $\{\mathbf{z}_{\ell,j}^{(\text{orig})} \}_{\ell,j=1}^{L,\, J} \gets f_\theta^{\text{LVLM}}(\mathbf{I}, \mathbf{q})$
\STATE $\{\mathbf{z}_{\ell,j}^{(\text{masked})} \}_{\ell,j=1}^{L,\, J} \gets f_\theta^{\text{LVLM}}(\tilde{\mathbf{I}}, \mathbf{q})$

\vspace{2pt}
\STATE \textcolor{myblue}{\textbf{\textit{Step 3: Subspace Alignment}}}
\FOR{$\ell \in \mathcal{L}$}
    \STATE $\boldsymbol{\delta}_{\ell,j} \gets \frac{1}{J}\sum_{j=1}^{J} \left(\mathbf{z}_{\ell,j}^{(\text{masked})} - \mathbf{z}_{\ell,j}^{(\text{orig})}\right)$
    \STATE $s_{k,\ell} \gets \left\| \boldsymbol{\delta}_{\ell,j}^\top \mathbf{V}_r^{(k)} \right\|_2 \quad \forall k$
    \STATE $\alpha_{k,\ell} \gets \frac{\exp(s_{k,\ell})}{\sum_{k'} \exp(s_{k',\ell})} \quad \forall k$
    \STATE $\gamma_k \gets \sum_{\ell \in \mathcal{L}} \alpha_{k,\ell} \quad \forall k$
\ENDFOR

\vspace{2pt}
\STATE \textcolor{myblue}{\textbf{\textit{Step 4: Adaptive Projection}}}
\STATE $\beta_k \gets \frac{\exp\left(\gamma_k / \tau\right)}{\sum_{k'} \exp\left(\gamma_{k'} / \tau\right)} \quad \forall k$
\STATE $\mathbf{P} \gets \sum_{k=1}^K \beta_k \mathbf{V}_r^{(k)} \left(\mathbf{V}_r^{(k)}\right)^\top$

\vspace{2pt}
\STATE \textcolor{myblue}{\textbf{\textit{Step 5: Project Hidden States On-the-Fly}}}
\STATE $\mathbf{z}_{\ell,j}^{(\text{edited})} \gets (\mathbf{I}_d - \mathbf{P}) \mathbf{z}_{\ell,j}^{(\text{orig})} \quad \forall \ell \in \mathcal{L},\ j \in \{1,\dots,J\}$

\vspace{2pt}
\STATE \textcolor{myblue}{\textbf{\textit{Step 6: Decoding from Edited States}}}
\STATE $y \gets \texttt{Decode}(\{ \mathbf{z}_{\ell,j}^{(\text{edited})} \})$

\vspace{2pt}
\RETURN $y$ \COMMENT{Final textual output (e.g., caption or answer)}
\end{algorithmic}

\end{algorithm}

Our proposed test-time strategy mitigates hallucinations in LVLMs by dynamically editing the model's internal activations using the subspace set \( \mathcal{V} \) constructed in the offline phase. For each test instance, we estimate its hallucination tendency and suppress activation components aligned with hallucination-inducing subspaces. {Unlike approaches that apply a fixed corrective signal to every input, our method treats each test sample as carrying its own latent mixture over hallucination modes, and infers this mixture on-the-fly through a lightweight perturbation-and-projection probe described below.}

Given a test image \( \mathbf{I} \in \mathbb{R}^{H \times W \times C} \), we first construct a masked variant by zeroing out 70\% of its semantically salient regions:
\begin{equation}
\tilde{\mathbf{I}} = \mathbf{I} \odot \mathbf{M},
\end{equation}
where \( \mathbf{M} \in \{0,1\}^{H \times W \times C} \) is a binary mask and \( \odot \) denotes element-wise multiplication. 

We then query the LVLM with both the original image \( \mathbf{I} \) and the masked image \( \tilde{\mathbf{I}} \), using the same textual prompt \( \mathbf{q} \). From a set of selected transformer layers \( \mathcal{L} \), we extract hidden representations and compute their difference to probe hallucination-sensitive behavior:
\begin{equation}
\boldsymbol{\delta}_{\ell,j} = \frac{1}{J} \sum_{j=1}^{J} \left( \mathbf{z}_{\ell,j}^{(\text{masked})} - \mathbf{z}_{\ell,j}^{(\text{orig})} \right) \in \mathbb{R}^d.
\end{equation}

Each difference vector \( \boldsymbol{\delta}_{\ell,j} \) is projected into the \( K \) pre-computed low-rank hallucination subspaces \( \{ \mathbf{V}_r^{(k)} \}_{k=1}^K \), where \( \mathbf{V}_r^{(k)} \in \mathbb{R}^{d \times r} \) contains orthonormal basis vectors for hallucination mode \( k \). The projection magnitude serves as an alignment score, $s_{k,\ell} = \left\| \boldsymbol{\delta}_{\ell,j}^\top \mathbf{V}_r^{(k)} \right\|_2.$ {Intuitively, a large \( s_{k,\ell} \) indicates that the visual ablation moves the representation strongly along directions previously identified with hallucination mode \( k \), suggesting that the test sample is particularly susceptible to that mode and should be corrected more aggressively along it.}

These scores are normalized per layer using softmax:
\begin{equation}
\alpha_{k,\ell} = \frac{\exp(s_{k,\ell})}{\sum_{k'=1}^K \exp(s_{k',\ell})}.
\end{equation}

Next, we aggregate alignment scores across layers to obtain a global importance score for each subspace, $\gamma_k = \sum_{\ell \in \mathcal{L}} \alpha_{k,\ell}$.

Then, a temperature-controlled softmax is then applied to derive adaptive weights:
\begin{equation}
\beta_k = \frac{\exp(\gamma_k / \tau)}{\sum_{k'=1}^K \exp(\gamma_{k'} / \tau)},
\end{equation}
where \( \tau > 0 \) is a temperature hyperparameter. Using these weights, we construct a sample-specific projection matrix:
\begin{equation}
\mathbf{P} = \sum_{k=1}^K \beta_k \mathbf{V}_r^{(k)} {\mathbf{V}_r^{(k)}}^\top.
\end{equation}

The operator \( \mathbf{P} \) inherits several useful properties from this construction. Since each term \( \mathbf{V}_r^{(k)} {\mathbf{V}_r^{(k)}}^\top \) is a symmetric positive semi-definite (PSD) orthogonal projector and the weights satisfy \( \beta_k \ge 0 \) with \( \sum_k \beta_k = 1 \), \( \mathbf{P} \) is itself symmetric PSD with spectral norm
\begin{equation}
\|\mathbf{P}\|_2 \le \max_k \|\mathbf{V}_r^{(k)} {\mathbf{V}_r^{(k)}}^\top\|_2 = 1,
\end{equation}
and the eigenvalues of \( \mathbf{I}_d - \mathbf{P} \) lie in \( [0, 1] \). Consequently, \( \mathbf{I}_d - \mathbf{P} \) is a (generalised) contraction that never amplifies any component of \( \mathbf{z}_{\ell,j} \), which is important for preserving generative fluency. Note that \( \mathbf{P} \) is in general \emph{not} idempotent: it reduces to a true orthogonal projection only in the degenerate cases where (i) the subspaces \( \{\mathbf{V}_r^{(k)}\}_{k=1}^K \) are mutually orthogonal, or (ii) the weights concentrate on a single mode, \( \beta_{k^\star} = 1 \).

Finally, we intervene in the forward pass of the LVLM by editing activations in-place. At each selected layer \( \ell \in \mathcal{L} \), token-wise hidden states are projected away from hallucination-prone subspaces:
\begin{equation}
\mathbf{z}_{\ell,j}^{\text{(edited)}} = (\mathbf{I}_d - \mathbf{P}) \mathbf{z}_{\ell,j}, \quad \text{for } j = 0, 1, \dots, J-1.
\end{equation}

This instance-specific projection dynamically suppresses hallucination-aligned components without requiring any parameter updates, thereby improving factual consistency while preserving the model's generative fluency. {Geometrically, \( \mathbf{I}_d - \mathbf{P} \) defines an oblique projection that removes only the component of each hidden state lying within the sample-specific mixture of hallucination subspaces, while leaving its orthogonal complement—containing the bulk of task-relevant semantic content—unchanged. Because \( \mathbf{P} \) is constructed entirely from pre-computed orthonormal bases and a handful of scalar weights, the intervention adds negligible computational overhead beyond a single additional forward pass on the masked image, and integrates seamlessly with any decoding scheme (greedy, beam, or sampling-based) without modifying the underlying model weights.}

\vspace{0.2cm}

\section{Experiments}
We evaluate our proposed method on two established hallucination detection benchmarks, CHAIR~\cite{rohrbach2018object} and POPE~\cite{li2023evaluating}—using three representative large vision-language models (LVLMs): LLaVA-1.5~\cite{liu2023visual}, MiniGPT-4~\cite{zhu2023minigpt}, and mPLUG-Owl2~\cite{ye2024mplug}. We compare our approach against a range of decoding-, tuning-, and projection-based baselines. Comprehensive ablation studies and qualitative analyses are also conducted to assess the impact of each design choice.

\subsection{Datasets}

\noindent
We evaluate our method on four benchmark datasets widely used to assess hallucination and visual grounding in large vision-language models: {CHAIR}~\cite{rohrbach2018object} and {POPE}~\cite{li2023evaluating}. These benchmarks collectively examine both object-level hallucination and multimodal consistency across image captioning and visual question answering tasks.

\paragraph{CHAIR.} CHAIR focuses on object hallucination in image captions by verifying whether the mentioned objects are visually grounded in the image. It reports two metrics: {CHAIR}$_S$, the percentage of captions containing hallucinated objects, and {CHAIR}$_I$, the proportion of hallucinated object tokens among all generated object mentions—lower values indicate better grounding. We also report BLEU to assess caption fluency. Following standard protocol~\cite{yang2025nullu}, we prompt each model with: \textit{``Please describe this image in detail.''}

\paragraph{POPE.} POPE evaluates hallucination through yes/no questions about object presence in images. It comprises three query types—\textit{random}, \textit{frequent}, and \textit{adversarial}—to probe model robustness under varying difficulty levels. In addition, we adopt the {Offline POPE (OPOPE)} variant~\cite{li2023evaluating}, which analyzes hallucination by checking whether non-existent objects appear in generated captions, rather than in direct answers to object queries.

\subsection{Baselines and Evaluation Setup}
We compare our method against a diverse set of recent hallucination mitigation approaches, including decoding-based, projection-based, and tuning-based techniques. Specifically, we evaluate against {DoLa}~\cite{chuang2023dola}, {OPERA}~\cite{huang2024opera}, {VCD}~\cite{leng2024mitigating}, {Woodpecker}~\cite{yin2024woodpecker}, {LURE}~\cite{zhou2023analyzing}, {HALC}~\cite{chen2024halc}, and {Nullu}~\cite{yang2025nullu}. We also include standard decoding strategies (\textit{Greedy}, \textit{Beam Search}) and tuning-based baselines such as LURE and MiniGPT-4 fine-tuned variants.

Our method is applied to three strong LVLM backbones: {LLaVA-1.5}~\cite{liu2023visual}, {MiniGPT-4}~\cite{zhu2023minigpt}, and {mPLUG-Owl2}~\cite{ye2024mplug}, all evaluated without any model fine-tuning. Following prior work~\cite{yang2025nullu}, we treat hallucination mitigation as a test-time operation and use consistent prompts across models for fair comparison. While our approach builds upon Nullu~\cite{yang2025nullu}, it differs by learning multiple HalluSpaces through clustering and applying adaptive, sample-specific null space projections.

\subsection{Implementation Details}
We apply our method without any fine-tuning to three pretrained LVLM backbones: LLaVA-1.5~\cite{liu2023visual}, MiniGPT-4~\cite{zhu2023minigpt}, and mPLUG-Owl2~\cite{ye2024mplug}. For each model and benchmark, we vary the number of HalluSpace clusters and the dimensionality of the projection bases. On the CHAIR benchmark, we use 5 clusters and 32 basis vectors for mPLUG-Owl2, and 11 clusters with 8 basis vectors for MiniGPT-4. On POPE, we use 6 clusters for LLaVA-1.5 and 11 clusters for MiniGPT-4. These settings were selected based on preliminary experiments balancing performance and computational efficiency. All evaluations are conducted on the MSCOCO validation split, and we report the mean and standard deviation across ten independent runs to account for variance introduced by clustering.

\begin{table*}[t]
\centering
\renewcommand{\arraystretch}{1.15}
\resizebox{1\textwidth}{!}{%
\begin{tabular}{l|ccc|ccc|ccc}
\toprule
\textbf{Method} & \multicolumn{3}{c|}{\textbf{LLaVA-1.5}} & \multicolumn{3}{c|}{\textbf{MiniGPT-4}} & \multicolumn{3}{c}{\textbf{mPLUG-Owl2}} \\
 & \textbf{CHAIR$_S \downarrow$} & \textbf{CHAIR$_I \downarrow$} & \textbf{BLEU$\uparrow$}
 & \textbf{CHAIR$_S \downarrow$} & \textbf{CHAIR$_I \downarrow$} & BLEU$\uparrow$
 & \textbf{CHAIR$_S \downarrow$} & \textbf{CHAIR$_I \downarrow$} & \textbf{BLEU$\uparrow$} \\
\midrule
Greedy & 20.40$_{\pm2.80}$ & 7.08$_{\pm0.33}$ & 15.72$_{\pm0.10}$ 
       & 32.40$_{\pm2.20}$ & 12.20$_{\pm0.42}$ & 14.57$_{\pm0.11}$ 
       & 22.90$_{\pm0.90}$ & 8.62$_{\pm0.11}$ & 15.01$_{\pm0.24}$ \\
Beam Search~\cite{freitag2017beam} 
       & 19.50$_{\pm2.30}$ & 6.84$_{\pm0.79}$ & 15.99$_{\pm0.14}$ 
       & 30.10$_{\pm0.30}$ & 11.87$_{\pm0.37}$ & 15.35$_{\pm0.24}$ 
       & 20.30$_{\pm0.70}$ & 7.62$_{\pm0.19}$ & 15.43$_{\pm0.05}$ \\
DoLa~\cite{chuang2024dola}
       & 20.20$_{\pm2.80}$ & 6.75$_{\pm0.54}$ & 15.68$_{\pm0.10}$ 
       & 31.90$_{\pm3.30}$ & 12.15$_{\pm0.89}$ & 14.54$_{\pm0.12}$ 
       & 22.40$_{\pm1.80}$ & 8.36$_{\pm0.04}$ & 15.13$_{\pm0.21}$ \\
OPERA~\cite{huang2024opera}
       & 17.50$_{\pm0.50}$ & 6.07$_{\pm0.32}$ & 16.02$_{\pm0.02}$ 
       & 29.70$_{\pm0.30}$ & 11.96$_{\pm0.29}$ & 14.82$_{\pm0.05}$ 
       & 20.07$_{\pm2.07}$ & 7.18$_{\pm0.39}$ & 15.41$_{\pm0.12}$ \\
VCD~\cite{leng2024mitigatingvcd} 
       & 20.30$_{\pm1.10}$ & 7.28$_{\pm0.10}$ & 14.53$_{\pm0.01}$ 
       & 29.00$_{\pm2.80}$ & 12.64$_{\pm1.19}$ & 14.42$_{\pm0.01}$ 
       & 22.80$_{\pm0.80}$ & 8.68$_{\pm0.17}$ & 15.14$_{\pm0.13}$ \\
Woodpecker~\cite{yin2024woodpecker}
       & 23.85$_{\pm4.62}$ & 7.50$_{\pm0.01}$ & 17.05$_{\pm0.00}$ 
       & 28.87$_{\pm2.20}$ & 10.20$_{\pm0.85}$ & 15.30$_{\pm0.01}$ 
       & 26.33$_{\pm1.98}$ & 8.43$_{\pm0.80}$ & 16.43$_{\pm0.00}$ \\
LURE~\cite{zhou2024analyzing}
       & 19.48$_{\pm2.35}$ & 6.50$_{\pm0.38}$ & 15.97$_{\pm0.01}$ 
       & 27.88$_{\pm2.25}$ & 10.20$_{\pm0.85}$ & 15.03$_{\pm0.01}$ 
       & 21.27$_{\pm0.06}$ & 7.67$_{\pm0.16}$ & 15.65$_{\pm0.15}$ \\
HALC~\cite{chen2024halc}
       & 16.90$_{\pm2.10}$ & 5.72$_{\pm0.55}$ & 16.02$_{\pm0.04}$ 
       & 25.20$_{\pm2.00}$ & 9.42$_{\pm0.41}$ & 14.91$_{\pm0.13}$ 
       & 18.80$_{\pm1.20}$ & 7.00$_{\pm0.01}$ & 15.33$_{\pm0.24}$ \\

Nullu~\cite{yang2025nullu} & {15.20$_{\pm0.60}$} & {5.30$_{\pm0.03}$} & {15.69$_{\pm0.04}$} 
              & {21.40$_{\pm1.00}$} & {8.99$_{\pm0.36}$} & {14.81$_{\pm0.06}$} 
              & {15.60$_{\pm1.20}$} & {5.77$_{\pm0.01}$} & {15.45$_{\pm0.01}$} \\
\midrule
\rowcolor{mygreen}
\textbf{Ours} & \textbf{14.60}{\tiny$\pm$0.35}  & \textbf{4.92}{\tiny$\pm$0.05} & 15.63{\tiny$\pm$0.01} 
              &\textbf{21.01}{\tiny$\pm$0.95} & \textbf{8.64}{\tiny$\pm$0.22} & 14.87{\tiny$\pm$0.03}
              & \textbf{15.21}{\tiny$\pm$1.01} & \textbf{5.46}{\tiny$\pm$0.01} & 15.69{\tiny$\pm$0.02} \\
            
\bottomrule
\end{tabular}
}
\vspace{3mm}
\caption{Comparison of different methods on CHAIR$_S$, CHAIR$_I$, and BLEU metrics across LLaVA-1.5, MiniGPT-4, and mPLUG-Owl2.}
\label{tab:results}
\end{table*}

\subsection{Results on CHAIR}

Table~\ref{tab:results} presents results on the CHAIR benchmark, using CHAIR$_S$, CHAIR$_I$, and BLEU as metrics. Across all three base models, our method consistently reduces both sentence-level and image-level hallucination rates compared to all baselines, including strong constrained decoding methods (HALC, DoLa) and null-space projection (Nullu). On LLaVA-1.5, we reduce CHAIR$_I$ from 5.30 (Nullu) to 4.92, while maintaining comparable BLEU. The trend holds for MiniGPT-4 and mPLUG-Owl2, where our method either matches or slightly improves BLEU, while reducing hallucination errors. Notably, methods such as DoLa and HALC, despite their sophisticated decoding strategies, exhibit a tendency to trade generation fluency for hallucination suppression, whereas our approach avoids this tension by operating directly in the visual feature space rather than intervening at the token probability level. The consistent gains across architecturally diverse base models — spanning Q-Former-based, linear-projection-based, and modality-collaboration-based designs — further underscore the generality of our sample-specific projection mechanism, which adapts to the idiosyncratic grounding failures of each model rather than applying a fixed correction. This demonstrates that our adaptive, sample-specific projection reduces hallucination rates more effectively than all baselines, while maintaining competitive BLEU scores, validating the effectiveness of targeted visual attribution constraints as a principled alternative to post-hoc decoding interventions.


\begin{table*}[t]
\centering
\resizebox{\textwidth}{!}{%
\setlength{\tabcolsep}{4pt}
\renewcommand{\arraystretch}{1.3}
\begin{tabular}{l|ccc|ccc|ccc}
\hline
\multirow{2}{*}{\textbf{Method}} & \multicolumn{3}{c|}{\textbf{LLaVA-1.5}} & \multicolumn{3}{c|}{\textbf{MiniGPT-4}} & \multicolumn{3}{c}{\textbf{mPLUG-Owl2}} \\
& \textbf{Accuracy↑} & \textbf{Precision↑} & \textbf{F score↑} & \textbf{Accuracy↑} & \textbf{Precision↑} & \textbf{F score↑} & \textbf{Accuracy↑} & \textbf{Precision↑} & \textbf{F score↑} \\
\hline
Greedy & 79.14{\tiny$\pm$0.89} & 91.98{\tiny$\pm$0.82} & 90.45{\tiny$\pm$0.86} & 71.22{\tiny$\pm$1.27} & 93.72{\tiny$\pm$1.02} & 90.04{\tiny$\pm$1.23} & 76.46{\tiny$\pm$0.92} & 88.85{\tiny$\pm$1.15} & 87.29{\tiny$\pm$1.15} \\
Beam Search~\cite{freitag2017beam} & 79.41{\tiny$\pm$0.69} & 92.52{\tiny$\pm$0.55} & 90.96{\tiny$\pm$0.59} & 71.65{\tiny$\pm$1.15} & 94.70{\tiny$\pm$0.60} & 90.97{\tiny$\pm$0.85} & 76.76{\tiny$\pm$1.02} & 90.28{\tiny$\pm$0.80} & 88.56{\tiny$\pm$0.87} \\
DoLa~\cite{chuang2024dola} & 78.98{\tiny$\pm$0.56} & 91.66{\tiny$\pm$0.81} & 90.15{\tiny$\pm$0.79} & 71.28{\tiny$\pm$1.15} & 93.92{\tiny$\pm$0.83} & 90.22{\tiny$\pm$1.04} & 76.07{\tiny$\pm$1.09} & 88.54{\tiny$\pm$1.25} & 86.95{\tiny$\pm$1.27} \\
OPERA~\cite{huang2024opera} & 79.29{\tiny$\pm$0.32} & 92.25{\tiny$\pm$0.07} & 90.71{\tiny$\pm$0.11} & 70.48{\tiny$\pm$1.63} & 94.41{\tiny$\pm$1.11} & 90.66{\tiny$\pm$1.42} & 75.49{\tiny$\pm$1.29} & 91.23{\tiny$\pm$1.06} & 89.11{\tiny$\pm$1.17} \\
VCD~\cite{leng2024mitigatingvcd} & 78.01{\tiny$\pm$0.75} & 91.33{\tiny$\pm$0.88} & 89.69{\tiny$\pm$0.89} & 70.83{\tiny$\pm$1.83} & 92.31{\tiny$\pm$0.88} & 88.76{\tiny$\pm$1.29} & 75.49{\tiny$\pm$1.27} & 88.75{\tiny$\pm$1.56} & 87.02{\tiny$\pm$1.57} \\
HALC~\cite{chen2024halc} & 77.87{\tiny$\pm$0.22} & 93.17{\tiny$\pm$0.39} & 91.25{\tiny$\pm$0.38} & 71.17{\tiny$\pm$0.89} & 94.88{\tiny$\pm$0.15} & 90.95{\tiny$\pm$0.42} & 74.93{\tiny$\pm$1.09} & 90.20{\tiny$\pm$0.90} & 88.12{\tiny$\pm$0.99} \\

Nullu~\cite{yang2025nullu} & 79.52{\tiny$\pm$0.04} & 93.46{\tiny$\pm$0.03} & 91.79{\tiny$\pm$0.04} & 71.92{\tiny$\pm$0.39} & 95.96{\tiny$\pm$0.65} & 92.07{\tiny$\pm$0.65} & 77.09{\tiny$\pm$1.37} & 92.83{\tiny$\pm$0.29} & 90.80{\tiny$\pm$0.52} \\
\hline
\rowcolor{mygreen}
\textbf{Ours} & \textbf{79.80}{\tiny$\pm$0.02} & \textbf{93.6}{\tiny$\pm$0.02} & \textbf{91.92}{\tiny$\pm$0.06} & \textbf{72.2}{\tiny$\pm$0.33} & \textbf{96.02}{\tiny$\pm$0.62} & \textbf{92.32}{\tiny$\pm$0.32} & \textbf{78.12}{\tiny$\pm$1.02} & \textbf{93.4}{\tiny$\pm$0.22} & \textbf{91.68}{\tiny$\pm$0.62} \\
\hline
\end{tabular}%
}
\vspace{3mm}
\caption{The OPOPE evaluation results on MSCOCO dataset of LVLMs with different methods for mitigating OH. Higher accuracy, precision, and F score indicate better performance.}
\label{tab:opope_results}
\end{table*}

\subsection{Results on POPE}

Table~\ref{tab:opope_results} reports results on POPE, which evaluates factual alignment in a question-answering setting. Our method achieves the best performance across all three metrics on mPLUG-Owl2, improving F-score from 90.80 (Nullu) to 91.60, along with consistent gains in accuracy and precision. For LLaVA-1.5 and MiniGPT-4, our method yields slight improvements in accuracy but does not surpass Nullu in F-score or precision. This disparity across architectures is informative: Q-Former-based models such as MiniGPT-4 compress visual information into a fixed token bottleneck before it reaches the language model, which may limit the degree to which visual attribution constraints applied upstream can recover fine-grained grounding signals already discarded during encoding. In contrast, mPLUG-Owl2's modality collaboration design retains richer visual context throughout the decoding pipeline, providing a more receptive substrate for our sample-specific suppression to operate on. These results suggest that our sample-specific suppression mechanism is particularly effective for stronger base models such as mPLUG-Owl2, where it enhances factual alignment without compromising fluency. Furthermore, POPE's binary yes/no probing format places a premium on precise object existence judgments rather than open-ended generation, making it a particularly stringent test of grounding fidelity — and one where even marginal F-score improvements reflect a meaningful reduction in systematic object confabulation. Moreover, the improvements demonstrate the adaptability of our framework, showing that even modest gains can accumulate to meaningful robustness in challenging benchmarks, and point toward future work on architecture-aware visual attribution that tailors constraint strength to the grounding capacity of the underlying connector design.


\subsection{Ablation Study}
We evaluate the impact of cluster count and basis dimensionality on hallucination mitigation. Fewer clusters result in reduced specificity, while smaller bases fail to isolate fine-grained spurious features. Our method outperforms Nullu even with fewer clusters, due to per-sample adaptivity.

\begin{figure*}[t!]
    \centering
    \includegraphics[width=\linewidth]{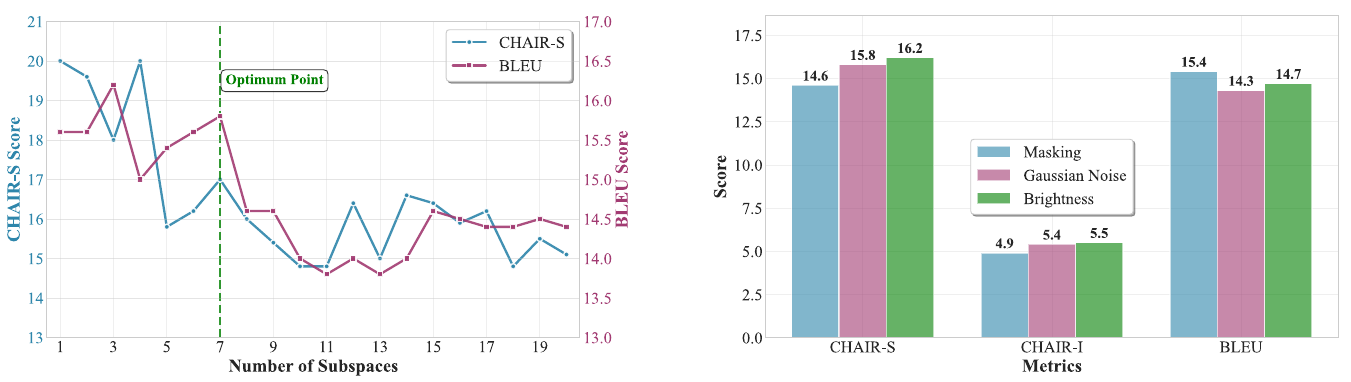}
    \vspace{3mm}
    \caption{ \textbf{(left)} The impact of the number of subspaces in the LLaVA-1.5 model within our proposed method. \textbf{(right)} We assess different perturbation strategies—masking, Gaussian, and blurring.
    }
    \label{fig:ablation_number_of_subspaces}
\end{figure*}

\vspace{0.2cm}
\noindent\textbf{Number of subspaces:} In this experiment (see Figure~\ref{fig:ablation_number_of_subspaces} \textbf{(left)}), we evaluate the effect of the number of subspaces on the performance of the LLaVA model. The optimal number of subspaces for each model is first determined on the COCO training set and then applied to the test set. As shown in the figure, we select 7 subspaces for LLaVA, as this configuration minimizes CHAIR\({}_S\) score while maximizing the BLEU score. Similarly, we choose 11 subspaces for MiniGPT-4 and 5 for mPLUG-Owl2 based on the same optimization criteria. These findings highlight that the optimal number of subspaces varies across models, underscoring the importance of tailoring the editing strategy to each LVLM. Moreover, the results confirm that increasing the number of subspaces beyond the optimal point does not necessarily yield further improvements and may degrade performance.

\noindent\textbf{Influence of image perturbations:} 
We evaluate the impact of different perturbation strategies—masking, Gaussian noise, and blurring—applied to test samples during test time in our hallucination mitigation method (see Figure~\ref{fig:ablation_number_of_subspaces} \textbf{(right)}). As shown in the figure, the masking strategy yields the best performance. This suggests that masking provides a more effective signal for disentangling hallucination-prone directions compared to other perturbations, likely because it introduces structured information removal that forces the model to rely on genuinely grounded visual features rather than spurious local statistics. Gaussian noise and blurring, by contrast, preserve the global spatial layout of the image while corrupting fine-grained content, which may insufficiently disturb the hallucination-inducing subspace and thus yield weaker contrastive signals for null-space estimation. These findings are consistent with prior work on contrastive visual decoding, where the quality of the degraded counterpart directly governs the fidelity of the suppression signal. Overall, the results highlight the importance of carefully selecting perturbation strategies when designing adaptive editing methods.


\noindent\textbf{Number of basis vectors:} This experiment (see Table \ref{tab:3}) evaluates the impact of the number of basis vectors in the LLaVA. As the number of basis vectors increases beyond 4, we observe a decrease in CHAIR$_{S}$ and CHAIR$_{I}$ scores, while the BLEU score decreases. This trend is undesirable, as a lower BLEU score in this context indicates that the LVLM model is generating responses that are less grounded in the image content. Intuitively, a larger basis spans a broader subspace of the visual feature space, which risks projecting out not only hallucination-prone directions but also semantically meaningful components that are essential for faithful image description. This over-suppression effect suggests that the null-space estimation becomes less discriminative as its dimensionality grows, collapsing signal that should be preserved alongside the noise it targets. The sweet spot at 4 basis vectors therefore reflects a balance between sufficiently constraining hallucination-inducing directions and retaining the residual visual information needed for coherent, grounded generation. As a result, we choose the number of basis vectors 4 for the LLaVA model.


\begin{table}[]
\centering
\scalebox{1}{
\begin{tabular}{l|ccccccc}
\hline
\multirow{2}{*}{Metric} & \multicolumn{7}{c}{Number of basis} \\
 & 1 & 2 & 4 & 8 & 16 & 32 & 64 \\ \hline
 \textbf{CHAIR$_S \downarrow$}& 16.2 & 15.2 & 14.6 & 14.5 & 13.2 & 12.5 & 11.3 \\
\textbf{CHAIR$_I \downarrow$}& 6.1 & 5.3 & 4.9 & 4.8 & 4.2 & 4.0 & 3.9 \\
\textbf{BLEU $\uparrow$}& 15.6 & 15.5 & 15.6 & 14.2 & 13.4 & 12.9 & 12.7 \\ \hline
\end{tabular}}
\vspace{3mm}
\caption{The influence of the number of subspaces in our method applied to the LLaVA-1.5 model.}
\label{tab:3}

\end{table}

\section{Conclusion}

We have presented a novel training-free, test-time adaptation method for mitigating hallucinations in large vision-language models. By modeling hallucinations with multiple low-rank subspaces derived from clustered hallucination-truthful feature pairs, our approach captures the diverse and sample-specific nature of hallucination patterns. Unlike existing methods, our framework dynamically adapts the hallucination suppression subspace for each test input, allowing for fine-grained, input-dependent corrections without permanently altering the model. Our extensive evaluation on six benchmarks and across four LVLM families confirms that our method consistently improves hallucination robustness while maintaining the original model’s integrity. This work offers a practical and effective solution toward more reliable LVLM deployments in real-world applications.

\bibliography{egbib}
\end{document}

%% file: math_commands.tex
\usepackage{amsmath,amsfonts,bm}

\def\eqref#1{equation~\ref{#1}}

\def\1{\bm{1}}

\DeclareMathAlphabet{\mathsfit}{\encodingdefault}{\sfdefault}{m}{sl}
\SetMathAlphabet{\mathsfit}{bold}{\encodingdefault}{\sfdefault}{bx}{n}

